\documentclass[10pt,twocolumn,letterpaper]{article}

\usepackage[final]{cvpr}              

\usepackage{colortbl}
\usepackage{bm}
\usepackage{amsmath}
\usepackage{MnSymbol}
\usepackage{multirow}
\usepackage{makecell}
\usepackage{booktabs}
\usepackage{nicefrac}
\usepackage{animate}
\definecolor{stefan}{rgb}{0.,0.33.,0.0}

\definecolor{hanzhi}{rgb}{0.08,0.33,0.6}

\definecolor{boyang}{rgb}{1.0,0.5,0.0}

\definecolor{cvprblue}{rgb}{0.21,0.49,0.74}
\definecolor{method}{rgb}{1.0,0.53,0.0}
\definecolor{author}{rgb}{0.2,0.1,0.6}

\usepackage[pagebackref,breaklinks,colorlinks,citecolor=cvprblue,linkcolor=cvprblue]{hyperref}
\newcommand{\authorhref}[3][author]{\href{#2}{\color{#1}{#3}}}

\def\confName{CVPR}
\def\confYear{2025}

\newcommand{\methodname}{\textit{VidBot}\xspace}

\title{\textcolor{method}{\methodname}: Learning Generalizable 3D Actions from In-the-Wild 2D Human \\ Videos for Zero-Shot Robotic Manipulation}

\author{%
  \authorhref{https://hanzhic.github.io/}{Hanzhi Chen}$^{1}$ \quad    
  \authorhref{https://boysun045.github.io/boysun-website/}{Boyang Sun}$^{2}$ \quad     
  \authorhref{https://dipan-zhang.github.io/}{Anran Zhang}$^{1}$ \quad  
  \authorhref{https://scholar.google.com/citations?user=YYH0BjEAAAAJ}{Marc Pollefeys}$^{2,3}$ \quad   
  \authorhref{https://scholar.google.ch/citations?user=SmGQ48gAAAAJ}{Stefan Leutenegger}$^{1, 2}$ 
  \\ 
  $^{1}$ Technical University of Munich \quad
  $^{2}$ ETH Zürich \quad
  $^{3}$ Microsoft \quad
  \\
}

\begin{document}

\twocolumn[{
\renewcommand\twocolumn[1][]{#1}%
\maketitle
\begin{center}
    \centering
    \vspace{-25pt}
    \includegraphics[width=1.0\textwidth]{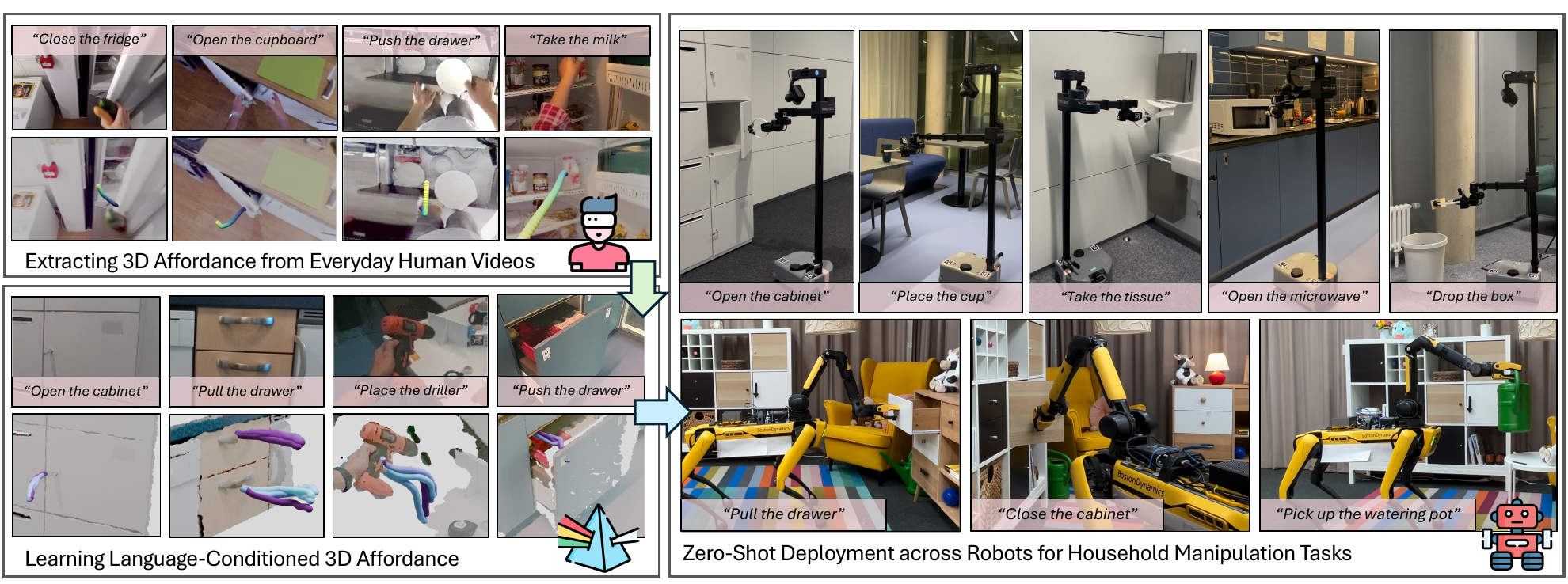}   
    \vspace{-20pt}
    \captionsetup{type=figure}\caption{\small{We present \textbf{\textcolor{method}{\methodname}}, a framework to learn interactions from in-the-wild RGB-only human videos. Our affordance model can be deployed across robots for daily manipulation tasks. Project website: \url{https://hanzhic.github.io/vidbot-project/}}}
    \label{fig:static}
\end{center}
}]

\begin{abstract}
\label{sec:abs}

Future robots are envisioned as versatile systems capable of performing a variety of household tasks. The big question remains, how can we bridge the embodiment gap while minimizing physical robot learning, which fundamentally does not scale well. We argue that learning from in-the-wild human videos offers a promising solution for robotic manipulation tasks, as vast amounts of relevant data already exist on the internet. In this work, we present \methodname, a framework enabling zero-shot robotic manipulation using learned 3D affordance from in-the-wild monocular RGB-only human videos. \methodname leverages a pipeline to extract explicit representations from them, namely 3D hand trajectories from videos, combining a depth foundation model with structure-from-motion techniques to reconstruct temporally consistent, metric-scale 3D affordance representations agnostic to embodiments. We introduce a coarse-to-fine affordance learning model that first identifies coarse actions from the pixel space and then generates fine-grained interaction trajectories with a diffusion model, conditioned on coarse actions and guided by test-time constraints for context-aware interaction planning, enabling substantial generalization to novel scenes and embodiments. Extensive experiments demonstrate the efficacy of \methodname, which significantly outperforms counterparts across 13 manipulation tasks in zero-shot settings and can be seamlessly deployed across robot systems in real-world environments. \methodname paves the way for leveraging everyday human videos to make robot learning more scalable. 

\end{abstract}

\section{Introduction}
\label{sec:intro}

Advancements in AI are accelerating the development of personalized devices, such as smart glasses that offer virtual guidance to users \cite{engel2023project, grauman2024ego, chen20243d, park2021review}. In the near future, robots will also become personalized systems, akin to smartphones or smart glasses, designed to provide physical assistance to humans. However, the diverse and novel forms of robotic embodiment pose a significant challenge for deploying AI to perform open-ended tasks in open-ended environments.

State-of-the-art approaches attempting to learn robot manipulation skills still rely heavily on human experts' teleoperated demonstrations, which are used to train robot policies under the Imitation Learning (IL) paradigm \cite{jang2022bc, johns2021coarse, zhang2018deep, pomerleau1988alvinn}. However, this process remains costly, time-consuming, and labor-intensive. While recent efforts have gathered large-scale robotic demonstrations for everyday manipulation tasks--such as Open X-Embodiment \cite{team2024octo} and DROID \cite{khazatsky2024droid}--scaling data collection remains challenging due to the combinatorial explosion of embodiments, tasks, and environments. We argue that human videos offer a promising scalable solution: there are already massive amounts of web videos capturing humans performing diverse tasks across various environments. Several previous approaches have explored human-to-robot skill transfer \cite{wang2023mimicplay, qin2022dexmv, bahl2022human, zhu2024vision, bahety2024screwmimic, shaw2023videodex, wang2024dexcap, xu2023xskill}. Nevertheless, they face certain limitations, such as requiring static cameras or scenes, depth sensors, MoCap systems, etc. 

These constraints often result in in-lab settings lacking diversity in scenes, illuminations, or viewpoints. One line of research has explored leveraging internet human videos with rich scene contexts to boost robot learning tasks, focusing on learning visual representations for visuomotor policies \cite{xiao2022masked, ma2022vip, nair2022r3m, radosavovic2023real}. However, one major limitation is the reliance on humans to collect task-specific teleoperated data in every new environment with every new embodiment to fine-tune the pre-trained model. More recently, works like \cite{bahl2023affordances} have progressed by explicitly extracting agent-agnostic interaction trajectories. Nevertheless, these extracted motions are simplified as 2D vectors in pixel space, limiting their direct deployment to robots. We argue that, beyond visual representations or pixel-level action cues limited to the 2D image plane, 3D affordance--specifically, the contact points and interaction trajectories with spatial awareness--is crucial for unifying different embodiments to interpret action from perception. However, extracting general 3D affordance data from everyday human videos remains a significant challenge, impeding robots from learning manipulation skills by watching humans.

In this work, we aim to enable zero-shot robot learning from a large amount of unlabeled everyday human videos, tackling two key questions: (1) How can 3D actionable knowledge be extracted from raw RGB-only human videos? (2) How can this extracted knowledge be reliably transferred to novel environments and new robot embodiments in a zero-shot manner? To answer the first question, we adopt a principled approach by exploiting Structure-from-Motion (SfM) for robot learning, developing a gradient-based optimization pipeline that extracts 3D hand trajectories from in-the-wild videos. Our pipeline combines learned monocular depth with geometric constraints, ensuring temporally consistent, metric-scale reconstructions. This allows for the recovery of contact points and smooth hand trajectories in 3D, serving as agent-agnostic 3D affordance representations.
We introduce a coarse-to-fine affordance learning framework for the second question to learn rich actions from the diverse extracted training data. At the coarse prediction stage, our affordance model identifies high-level actions from pixels, i.e., contact points and goal points, based on RGB-D observations and task instructions. In the fine prediction stage, we employ a diffusion model to generate fine-grained interaction trajectories conditioned on coarse-stage outputs and task observations.
Rather than relying solely on learned action priors from humans, we incorporate several differentiable cost functions as test-time sampling guidance \cite{ho2022video}. These cost functions guide the diffusion denoising process by perturbing outputs to satisfy test-time constraints during deployment. Differentiable objectives, such as multi-goal reaching and collision avoidance, capture the distribution of coarse outputs and leverage geometric cues from the scene context. 
The guidance terms improve the plausibility of interactions by accounting for the variance in scene contexts and robot morphology while providing intuitive heuristics for selecting the optimal plans. 

We conducted extensive experiments in both simulation and real-world settings to evaluate the effectiveness of our framework. Utilizing in-the-wild 2D human videos only, our 3D affordance model outperformed several baselines trained using simulator exploration or pre-trained with teleoperated demonstrations. Furthermore, we demonstrated our model's versatility in various downstream robot learning tasks, such as visual goal-reaching and exploration, showcasing rapid convergence to superior performance. 

\section{Related Work}
\label{sec:related}

\subsection{Visual Affordance Learning}
\textit{Affordance} centers around determining \textit{where} and \textit{how} an agent should interact with a given scene. One line of work regress affordance using manually annotated datasets \cite{do2018affordancenet, chuang2018learning, myers2015affordance, delitzas2024scenefun3d}. However, collecting affordance labels is highly costly. Hence, a more recent line of work addressed this challenge by deploying agents in simulated environments to explore effective interaction \cite{mo2021where2act, ning2024where2explore, wu2021vat, geng2023partmanip, chen2024funcgrasp}. 
Despite offering a data collection alternative without human intervention, these methods often suffer from the cost of obtaining diverse virtual assets.
In contrast, human videos have gained attention as a more general source of affordance priors. Several approaches \cite{liu2022joint, goyal2022human, bahl2023affordances, nagarajan2019grounded} predict per-pixel affordance scores by leveraging hand-object contact labels from human videos. However, these pipelines usually only identify contact regions or model interaction actions within the image plane, lacking spatial awareness.
More recent works \cite{yuan2024general, bharadhwaj2024track2act} attempted to address this limitation by utilizing flows as spatially-aware affordance representations. However, these approaches require either goal images or initial contact regions given at test time. In contrast, our affordance model eliminates these requirements and directly infers contact points and interaction trajectories in 3D, as learned from in-the-wild RGB-only human videos.

\subsection{Robot Learning from Humans}
Previous works have explored utilizing human videos to aid robot learning tasks. One approach involves learning visual representations from human videos and using pre-trained visual encoders to train policy networks \cite{nair2022r3m, xiao2022masked, ma2022vip, bharadhwaj2024towards, radosavovic2023real, xu2024flow, wen2023anypoint}. Another line of research focuses on learning reward functions from human videos \cite{chang2024look, li2024ag2manip, bahl2022human, smith2019avid, xiong2021learning, chen2021learning, xu2023xskill, wang2023mimicplay, liao2024toward}. Additionally, some works use motion attributes extracted from videos, such as estimating 3D hand poses or tracking wrist trajectories \cite{qin2022dexmv, sivakumar2022robotic, wang2024dexcap, shaw2023videodex, ye2023learning, bharadhwaj2023zero, wang2023mimicplay, papagiannis2024r+}. However, these methods are typically restricted to in-lab setups and/or require further teleoperated demonstrations by human experts. \cite{bahl2023affordances} is the closest work to ours, which used everyday human videos to extract embodiment-agnostic actions. However, its inferred 2D pixel-level motions are oversimplified and ambiguous, limiting direct deployment to robots. In contrast, our framework leverages the same human video data as \cite{bahl2023affordances} for supervision but can predict affordance in 3D space, enabling zero-shot skill transfer to robots.

\subsection{Diffusion Models in Robotics}
Diffusion models are a powerful learning paradigm approximating complex data distributions through an iterative denoising process. Recently, they have achieved success across various generative modeling applications \cite{kingma2021variational, wang2024move, ho2020denoising, rombach2022high, ramesh2022hierarchical, ho2022video, chou2023diffusion, zhao2024michelangelo, rempe2023trace, gao2024diffcad}. In robotics, diffusion models have shown to be strong policy learning frameworks \cite{chi2023diffusionpolicy, janner2022diffuser, liang2023adaptdiffuser, ma2024hierarchical, xian2023chaineddiffuser, liang2024skilldiffuser, ajay2022conditional, kang2024efficient}. Diffusion Policy \cite{chi2023diffusionpolicy} introduced a general framework for generating multi-modal robot trajectories via a conditional denoising diffusion process. Diffuser \cite{janner2022diffuser} enhanced guided trajectory sampling by incorporating reward functions. Follow-up works \cite{xian2023chaineddiffuser, ma2024hierarchical, liang2024skilldiffuser} have proposed more factorized policy learning frameworks that allow diffusion models to generate smooth actions between key steps. However, these approaches focus on regressing highly limited in-domain teleoperation data with no modality or embodiment gaps during testing. In contrast, our approach learns policy from massive training data extracted from human videos. We introduce a coarse-to-fine affordance learning framework integrated with cost guidance to enhance generalization and test-time flexibility in novel scenes with new embodiments.

\section{Method}
\label{sec:method}

\subsection{Problem Definition} 
\label{subsec:problem}
We aim to learn a factorized affordance model $\mathbf{a} = \pi(\{\Tilde{\mathbf{I}}, \Tilde{\mathbf{D}}\}, l)$ from everyday human videos, where $\{\Tilde{\mathbf{I}}, \Tilde{\mathbf{D}}\}$ is an RGB-D frame (image $\Tilde{\mathbf{I}}$, depth $\Tilde{\mathbf{D}}$), and $l$ is language instruction. Note the depth frame can be obtained either from a depth sensor or a metric-depth foundation model \cite{bhat2023zoedepth, yang2024depth}. As the affordance representation is expected to be embodiment-agnostic, we formulate the final output affordance representation $\mathbf{a}$ as contact points $\mathbf{c}$ and interaction trajectories $\bm{\tau}$ following previous works \cite{bahl2023affordances, liu2022joint}, while extending this formulation into 3D space. Specifically, $\mathbf{a} = \{\mathbf{c}, \bm{\tau}\}$, where $\mathbf{c} \in \mathbb{R}^{N_{\text{c}} \times 3}, \bm{\tau} \in \mathbb{R}^{H \times 3}$. Here, $N_{\text{c}}$ is the number of contact points, and $H$ is the trajectory horizon. Note $\mathbf{a}$ is represented in the observation camera's frame.

\subsection{3D Affordance Acquisition from Human Videos}
We first design a pipeline to extract 3D hand trajectories from daily human videos recorded by a \textbf{moving} monocular camera, where each frame's pose and  are \textbf{unknown}.
Here, we introduce the key components in our pipeline.

\textbf{Data Preparation.} Given a video with color images $\{\hat{\mathbf{I}}_0, ...,  \hat{\mathbf{I}}_T\}$ and language description $l$, an SfM system \cite{schoenberger2016sfm} is first employed to estimate camera intrinsics $\mathbf{K}$, per-frame scale-unaware poses $\{\mathbf{T}_{\text{W} \text{C}_0}, ...,  \mathbf{T}_{\text{W} \text{C}_T}\}$ and sparse landmarks $\{_{\text{W}}\mathbf{l}_0, ...,  {_{\text{W}}}\mathbf{l}_{N_l}\}$ expressed in the world frame. We harness a metric-depth foundation model \cite{yang2024depth, bhat2023zoedepth, hu2024metric3dv2} to predict each frame's dense depth $\{\hat{\mathbf{D}}_0, ...,  \hat{\mathbf{D}}_T\}$. We further utilize a hand-object detection model \cite{shan2020understanding} and segmentation models \cite{zhang2022fine, kirillov2023segment} to acquire the masks of each frame's hand and in-contact object, i.e., $\{\mathbf{M}_0^\text{h}, ...,  \mathbf{M}_T^\text{h}\}$, $\{\mathbf{M}_0^\text{o}, ...,  \mathbf{M}_T^\text{o}\}$. With the hand masks provided, we further collect frames before $\mathbf{I}_0$ and their hand masks to obtain hand-less frames $\{\Tilde{\mathbf{I}}_0, ...,  \Tilde{\mathbf{I}}_T\}$ with a video inpainting model \cite{li2022towards}.

\textbf{Consistent Pose Optimization.} Our first objective is to correct the camera poses to the metric-space scale. To achieve this goal, we optimize a global scale $s_\text{g}$ for all frames by projecting the sparse landmarks to each image plane using camera intrinsics and its pose:
\begin{equation}
\begin{aligned}
\operatorname*{min}_{s_{\text{g}}} \hspace{-0.9mm} \sum_{i, j} \hspace{-0.9mm} \Tilde{\mathbf{M}}_i[\mathbf{{u}}_{ij}] \left\Vert \hat{\mathbf{D}}_i[\mathbf{{u}}_{ij}] - s_{\text{g}}\operatorname{d}(\mathbf{T}_{\text{W} \text{C}_i}^{-1} {_{\text{W}}\mathbf{l}_j}) \right\Vert_2^2,
\end{aligned}
\label{eqn:scale_opt}
\end{equation}
where $\Tilde{\mathbf{M}}_i= \neg(\mathbf{M}_i^\text{h} \bigcup \mathbf{M}_i^\text{o})$, denoting static regions, $\mathbf{{u}}_{ij}$ is the pixel coordinate of landmark $j$ in camera $i$ and $\operatorname{d}(\cdot)$ is the depth of the point. 
We then refine all frames' poses $\mathbf{T}_{\text{W} \text{C}_i} \in \mathcal{T}$ and scales $s_i \in \mathcal{S}$ to compensate SfM reconstruction errors due to dynamic hand-object motions and simultaneously make the predicted depth more consistent across views by optimizing the following term:
\begin{equation}
\begin{aligned}
  \operatorname*{min}_{\mathcal{T} \backslash \{\mathbf{T}_{\text{W} \text{C}_k} \} , \mathcal{S} \backslash \{s_{k}\}} \sum_{i\neq k} \sum_{\mathbf{u}_{i}}  \Tilde{\mathbf{M}}_{i}[\mathbf{u}_{i}] \Tilde{\mathbf{M}}_{k}[\mathbf{u}_{k}] \mathbf{E}[\mathbf{u}_{i}],
\end{aligned}
\label{eqn:pose_opt}
\end{equation}
where $k$ is the reference frame index yielding the highest co-visibility with the others, ${_{\text{C}_{i}}}\mathbf{X}_i^n$ is the back-projected points using the camera intrinsics $\mathbf{K}$ and the depth $\hat{\mathbf{D}}_i$.
  $\mathbf{E}[\mathbf{u}_{i}] = \left\Vert s_i^{-1}\mathbf{T}_{\text{C}_{k}\text{C}_{i}}{_{\text{C}_i}\mathbf{X}_i^n}[\mathbf{u}_{i}] - s_{k}^{-1} {_{\text{C}_k}\mathbf{X}_{k}^n}[\mathbf{u}_{k}]  \right\Vert_2^2$, 
  where $\mathbf{u}_{k}$ is the projective pixel correspondence to $\mathbf{u}_{i}$ computed with $s_i, s_k, \mathbf{K}, \hat{\mathbf{D}}_i$, and $\mathbf{T}_{\text{C}_{k}\text{C}_{i}}=\mathbf{T}_{\text{W} \text{C}_{k}}^{-1} \mathbf{T}_{\text{W} \text{C}_i}$. $s_k$ is fixed to $s_\text{g}$.  
  
\begin{figure}[t]
	\centering 
        \includegraphics[width=0.9\linewidth]{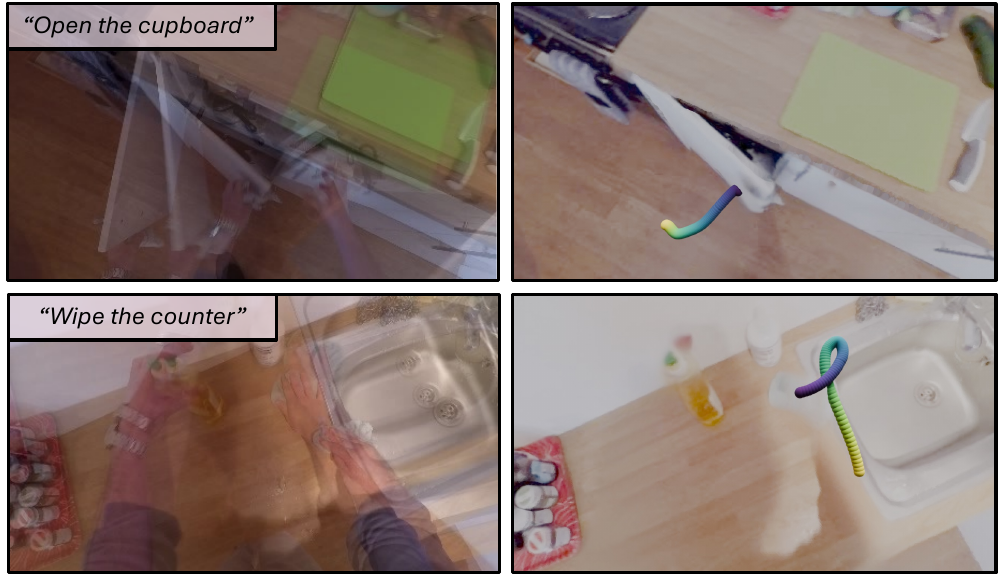} 
	\centering
        \vspace{-0.1 in}
	\caption{\small Example trajectories extracted from raw human videos.} 
         \label{fig:traj_demo}
        \vspace{-0.2 in}
\end{figure}

\textbf{Affordance Extraction.} 
We obtain each frame's hand center point and transform it to the first frame with the refined poses and scales to compute the interaction trajectory $\hat{\bm{\tau}}$. We downsample hand points uniformly in the first frame to acquire contact points $\hat{\mathbf{c}}$, and them from the last frame to acquire goal points $\hat{\mathbf{g}}$ to supervise the intermediate prediction of our affordance model. 
Language description $l$, inpainted color $\Tilde{\mathbf{I}}_0$ and its depth $\Tilde{\mathbf{D}}_0$ from \cite{yang2024depth}, together with the inpainted object image $\Tilde{\mathbf{I}}_0^{o}$ cropped using $\mathbf{M}_0^\text{o}$, are used as model inputs. We leverage the EpicKitchens-100 Videos dataset \cite{Damen2022RESCALING} and its SfM results provided by EpicFields \cite{EPICFields2023} to showcase the effectiveness of our pipeline. Fig. \ref{fig:traj_demo} exemplifies the extracted results.

\begin{figure*}[t]
	\centering 
        \vspace{-0.25in}
        \includegraphics[width=1.0\linewidth]{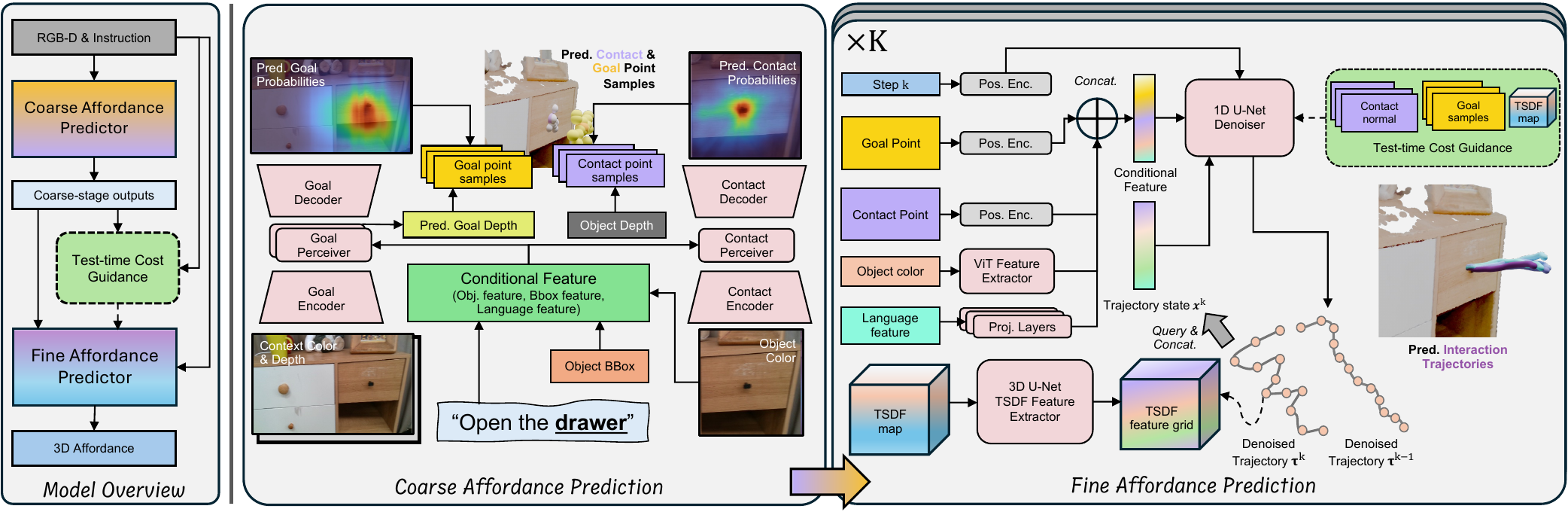} 
	\centering
        \vspace{-0.25in}
	\caption{Overview of our affordance learning model. The affordance model is factorized into a coarse stage and a fine stage. We parse high-level contact and goal configurations from task inputs in the coarse stage. Supp. Mat. provides more detailed illustration of conditional feature extraction process. In the fine stage, we utilize the coarse stage outputs to guide the fine-grained interaction trajectory generation process through conditioning and cost functions. The color represents the final cost value, with darker shades indicating lower costs. } 
        \label{fig:pipeline}
        \vspace{-0.2 in}
\end{figure*}

\subsection{Coarse-to-Fine Affordance Learning}
The overview of our affordance model is shown in Fig. \ref{fig:pipeline}. We design the model with two key considerations: (1) It should capture the action distribution conditioned on observation and instruction from massive in-the-wild human affordance data. (2) It should leverage contextual information during test time to mitigate the embodiment gap and potentially noisy prediction due to imperfect training data, thereby enhancing the quality of the generated affordance. 

To address the first aspect, we factorize the affordance model $\pi$ into a coarse model $\pi_{\text{c}}$ and a fine model $\pi_{\text{f}}$. In the coarse stage, $\pi_{\text{c}}$ performs high-level scene understanding to infer a set of goal points $\mathbf{g}$ and contact points $\mathbf{c}$ conditioned on the RGB-D frame $\{\Tilde{\mathbf{I}}, \Tilde{\mathbf{D}}\}$ and instruction $l$, i.e., $\{\mathbf{g}, \mathbf{c}\} = \pi_{\text{c}}(\{\Tilde{\mathbf{I}}, \Tilde{\mathbf{D}}\}, l), \hspace{1mm} \mathbf{a}_{\text{c}}=\{\mathbf{g}, \mathbf{c}\}$. Given the coarse-stage outputs together with the task inputs, $\pi_{\text{f}}$ plans fine-grained interaction trajectories at a low level with $\bm{\tau} = \pi_{\text{f}}(\{\Tilde{\mathbf{I}}, \Tilde{\mathbf{D}}\}, l, \bm{a}_{\text{c}})$. To achieve the second objective, we integrate multiple analytical cost functions for $\pi_{\text{f}}$, incorporating scene context and agent embodiment during the test time. These constraints guide the trajectory generation process, leading to more plausible and context-aware interaction trajectories.

The contact points $\mathbf{c}$ and the interaction trajectory $\bm{\tau}$ will be the final affordance outputs $\mathbf{a} = \{\mathbf{c}, \bm{\tau}\}$.

\subsubsection{Coarse Affordance Prediction} 
At this stage, the coarse affordance model is designed to extract macro actionable information from high-dimensional image space. To achieve this, we represent the coarse action points in pixel space by learning probabilities of the coarse affordance, along with their corresponding depth, where applicable. As illustrated in Fig. \ref{fig:pipeline}, we first obtain a cropped RGB-D image of the objects of interest using an off-the-shelf open-set object detector \cite{liu2023grounding, xiong2024efficientsam}. Our coarse model, $\pi_{\text{c}}$, consists of two networks: $\pi_{\text{c}}^{\text{goal}}$ and $\pi_{\text{c}}^{\text{cont}}$, which predict the goal and contact points, respectively.

For goal points prediction, the context color image and a depth image filled with the median depth of the object of interest is fed to $\pi_{\text{c}}^{\text{goal}}$ to obtain the goal heatmap activation with the exact resolution, together with the depth of the goal points, as the endpoint of an interaction trajectory tends to be distributed in free space.  Given the \textit{global context feature} $\mathbf{z}^{\text{goal}}$ encoded by the visual encoder of $\pi_{\text{c}}^{\text{goal}}$, we extract the object-centric embedding $\mathbf{z}_{\text{o}}^{\text{goal}}$ using RoI Pooling \cite{girshick2015fast}. Further, we acquire bounding box positional feature $\mathbf{z}_{b}^{\text{goal}}$ from a multi-layer perceptron (MLP) and the feature embedding $\mathbf{z}_{l}$ of the language instruction using a frozen CLIP model \cite{radford2021learning}. Given the conditional feature $\mathbf{z}_{\text{c}}^{\text{goal}} = \{ \mathbf{z}_{\text{o}}^{\text{goal}}, \mathbf{z}_{\text{b}}^{\text{goal}}, \mathbf{z}_{\text{l}}\}$, we leverage a Perceiver \cite{jaegle2021perceiver} module with several self-attention and cross-attention blocks, enabling the global context feature $\mathbf{z}^{\text{goal}}$ to attend to the conditional feature $\mathbf{z}_{\text{c}}^{\text{goal}}$. The fused \textit{global context feature} will be first passed to the transformer encoder and MLP layers to predict the goal depth and forwarded to a visual decoder to predict the per-pixel goal probabilities. 

The contact predictor follows a similar hourglass network architecture as the $\pi_{\text{c}}^{\text{goal}}$. However, we omit the prediction of contact points' depth as they tend to lie on the objects' surface with valid depth. Only the language feature $\mathbf{z}_{\text{l}}$ is fused into the \textit{object contact feature} $\mathbf{z}^{\text{cont}}$ extracted by the visual encoder of $\pi_{\text{c}}^{\text{cont}}$. 

Finally, using the camera intrinsics, sampled pixel coordinates from the predicted heatmaps, and their queried depth, we lift them to 3D to obtain the coarse affordance output, i.e., the goal points $\mathbf{g}$, and the contact points $\mathbf{c}$.

\subsubsection{Fine Affordance Prediction} 
The fine affordance model is conditioned to infer a fine-grained interaction trajectory guided by the contact and goal points. This stage is modeled as a conditional diffusion denoising process inspired by \cite{janner2022diffuser}.

\textbf{Diffusion models preliminaries.}
\label{subsec:diffusion}
Our fine affordance model follows the diffusion probabilistic model formulation \cite{ho2020denoising}. Such a formulation is modeled with a \textit{forward process} and a \textit{reverse process}. 

Given a sample $\bm{\tau}^0$ drawn from its underlying distribution $q(\bm{\tau})$,  the \textit{forward process} iteratively injects Gaussian noise in $K$ steps as a Markovian process. Such a process can be expressed as:
\begin{equation}
\begin{aligned}
q(\bm{\tau}^{k} | \bm{\tau}^{k-1} ) &= \mathcal{N}(\bm{\tau}^{k}; \sqrt{1 - \beta_{{k}}} \bm{\tau}^{k-1}, \beta_{{k}} \textbf{I}), \\
q(\bm{\tau}^{1:K} | \bm{\tau}^\text{0} ) &= \prod_{k=1}^{K} q(\bm{\tau}^{k} | \bm{\tau}^{k-1}),
\end{aligned}
\label{eqn:diffusion_forward}
\end{equation}
where $\beta_{{k}}$ is acquired from a pre-defined scheduler.

In the \textit{reverse process}, a denoising neural network $ \phi$ learns distribution $p_{\phi}(\bm{\tau}^{k-1} | \bm{\tau}^k)$ to gradually remove the noise so as to recover $\bm{\tau}^0$:
\begin{equation}
\begin{aligned}
p_{\phi}(\bm{\tau}^{k-1} | \bm{\tau}^{k} ) &= \mathcal{N}(\bm{\tau}^{k-1}; \bm{\mu}_{\phi}(\bm{\tau}^{k}, k), \bm{\Sigma}_k), \\
p_{\phi}(\bm{\tau}^{1:K}) &= p(\bm{\tau}_K)\prod_{k=1}^{K} p_{\phi}(\bm{\tau}^{k-1} | \bm{\tau}^{k} ), 
\end{aligned}
\label{eqn:diffusion_reverse}
\end{equation}
where $\bm{\mu}_{\phi}(\bm{\tau}^{k}, k)$ is from the denoising neural network, and $\bm{\Sigma}_k$ is from a fixed scheduler.

The desired diffusion state $\bm{\tau}$ is the interaction trajectory in our setting. The denoising network learns a conditional distribution as $p_{\phi}(\bm{\tau}^{k-1} | \bm{\tau}^{k}, \bm{o})$ conditioned on task observations $\bm{o}$ to be explained, i.e., $\bm{\mu}^k = \bm{\mu}_{\phi}(\bm{\tau}^{k}, k, \bm{o})$.

\textbf{Trajectory generation.} The fine affordance model $\pi_{\text{f}}$ is parameterized by a 1D U-Net similar to \cite{rempe2023trace}. Specifically, we first acquire the goal point $\bar{\mathbf{g}}$ and contact point $\bar{\mathbf{c}}$ from $\mathbf{g}$ and $\mathbf{c}$ with the highest predicted probabilities. Moreover, we obtain the object's feature embedding $\mathbf{z}^\text{fine}_{\text{o}}$ extracted by a vision transformer \cite{caron2021emerging}. Together with the language instruction embedding $\mathbf{z}_{l}$ extracted in the coarse stage, we establish the conditional embedding as $\bm{o} = \{\operatorname{PE}(\mathbf{g}), \operatorname{PE}(\mathbf{c}),  \operatorname{Proj}(\mathbf{z}_{l}), \mathbf{z}^\text{fine}_{\text{o}}\}$, $\operatorname{PE}$ is the positional encoding operator and $\operatorname{Proj}$ denotes blocks with a transformer encoder \cite{vaswani2017attention} and an MLP. To integrate spatial awareness into the inferred trajectory, we encode $C_{m}$-dimensional latent features from the voxelized TSDF map $\mathbf{U}$ acquired from RGB-D frame using a 3D U-Net \cite{peng2020convolutional} and compute the spatial feature $\mathbf{f}^k \in \mathbb{R}^{H \times C_{m}}$ for each waypoint from denoised trajectory $\bm{\tau}^k$ using trilinear interpolation. $\mathbf{f}^k$ and $\bm{\tau}^k$ are concatenated together as denoising inputs $\mathbf{x}^k = \{\bm{\tau}^k, \mathbf{f}^k \} $ fed to $\pi_{\text{f}}$. Instead of using noise-prediction, $\pi_{\text{f}}$ directly infers the unnoised trajectory $\bar{\bm{\tau}}^{0}$ in each step $k$ and uses it to compute $\bm{\mu}^k$ (\textit{c.f.} Supp.\ Mat. for details), i.e., $\bar{\bm{\tau}}^{0} = \pi_{\text{f}}(\mathbf{x}^k, \operatorname{PE}(k), \bm{o})$. 

\subsubsection{Cost-Guided Trajectory Generation} 
The inferred trajectory could become erroneous if the conditioned goal point $\bar{\mathbf{g}}$ has an offset. This is expected since $\pi_{\text{f}}$ essentially acts as a gap filler between the contact point $\bar{\mathbf{c}}$ and the goal point $\bar{\mathbf{g}}$. The optimal goal point for conditioning may not always be selected based on the predicted scores, and multiple goal points from the goal set $\mathbf{g}$ can yield more diverse and robust predictions. However, querying the affordance model multiple times by sampling different goal configurations is computationally inefficient. Therefore, we cast the multi-goal conditioning as a cost function to guide trajectory generation during test time. Specifically, we define the multi-goal cost function as:
\begin{equation}
\begin{aligned}
\mathcal{J}_{\text{goal}} &= \min\limits_{\mathbf{g}_{n} \in \mathbf{g}} {\left\Vert \mathbf{g}_{n} - \bar{\bm{\tau}}^{0}_{H}\right\Vert^2_2},
\end{aligned}
\label{eqn:guidance_goal}
\end{equation}
where $\bar{\bm{\tau}}^{0}_{H} $ denotes the trajectory endpoint and $\mathbf{g}_{n}$ is the $n$-th goal point.
We additionally formulate scene collision avoidance guidance and contact normal guidance as $\mathcal{J}_{\text{collide}}$ and $\mathcal{J}_{\text{normal}}$, where $h$ denotes the trajectory horizon index:
\begin{equation}
\begin{aligned}
\mathcal{J}_{\text{collide}} &= \frac{1}{H^{\prime}N_{\text{p}}} \sum_{h \neq 1, i} - \min\left(\mathbf{U}[\mathbf{p}_i + \bar{\bm{\tau}}^{0}_{h} -\bar{\bm{\tau}}^{0}_{1}], 0\right) , \\
\mathcal{J}_{\text{normal}} &=  \frac{1}{H^{\prime}} \sum_{h \neq 1} \min\limits_{s_{\text{n}} \in \{-1, 1\}} \left\Vert  \nicefrac{(\bar{\bm{\tau}}^{0}_{h} - \bar{\bm{\tau}}^{0}_{1})}{\left\Vert\bar{\bm{\tau}}^{0}_{h} - \bar{\bm{\tau}}^{0}_{1}\right\Vert_2} - s_{\text{n}}\mathbf{n} \right\Vert^2_2.
\end{aligned}
\label{eqn:guidance_others}
\end{equation}
We sample $N_{\text{p}}$ points from both the agent's hand model (e.g., robot gripper) and the object's surface (if it is portable) prior to interaction ($h=1$), where each one is denoted as $\mathbf{p}_i$, and query their values differentiably from the pre-computed TSDF map $\mathbf{U}$. The normal vector $\mathbf{n}$ is computed from the contact points. The trajectory starting point $\bar{\bm{\tau}}^{0}_{1}$ is not optimized during guidance, hence $H^{\prime}=H-1$. 
The final cost function $\mathcal{J}$ for test-time guidance is formulated as: 

\begin{equation}
\begin{aligned}
\mathcal{J} = 
\lambda_{\text{g}} \mathcal{J}_{\text{goal}} + 
\lambda_{\text{c}} \mathcal{J}_{\text{collide}} + 
\lambda_{\text{n}} \mathcal{J}_{\text{normal}},  
\end{aligned}
\label{eqn:guidance_sum}
\end{equation}
where $\lambda_{\text{g}}$, $\lambda_{\text{c}}$, $\lambda_{\text{n}}$ control the strength of each guidance term.

We adopt \textit{reconstruction guidance} from \cite{ho2022video, rempe2023trace}. At each denoising step, gradients of $\mathcal{J}$ w.r.t.\ $\bm{\tau}^{k}$  will adjust the unnoised predictions $\bar{\bm{\tau}}^{0}$ as $\bm{\tau}^{0}$:
\begin{equation}
\begin{aligned}
\bm{\tau}^{0} = \bar{\bm{\tau}}^{0} - \bm{\Sigma}_{k} \nabla_{\bm{\tau}^k} \mathcal{J}.
\end{aligned}
\label{eqn:reverse_update}
\end{equation}
Now we use $\bm{\tau}^{0}$ instead of $\bar{\bm{\tau}}^{0}$ to compute $\bm{\mu}^k$. 

Introducing test-time guidance into the trajectory generation process offers several advantages: 1) Trajectories can better capture the goal distribution without extensive forward passes through the fine affordance model. 2) The morphology of novel embodiments and the geometry of previously unseen objects can be accounted for, providing collision-free hand trajectories readily integrated into downstream whole-body planning. 3) The final cost value $\mathcal{J}$ for each trajectory is an informative criterion for the agent to select the optimal interaction plan.

\subsubsection{Affordance Models Training}
To train the coarse affordance model, i.e., $\pi_\text{c}^\text{goal}$ and $\pi_\text{c}^\text{cont}$, we project the extracted goal points $\hat{\mathbf{g}}$ and contact points $\hat{\mathbf{c}}$ to the image plane and obtain ground-truth probabilities by fitting a Gaussian mixture model to them, which results in $\hat{\mathbf{H}}_\text{g}$ and  $\hat{\mathbf{H}}_\text{c}$. The goal depth $\hat{{D}}_\text{g}$ is additionally regressed by $\pi_\text{c}^\text{goal}$, which is the median depth of the goal points. We include an auxiliary vector field regression loss $\mathcal{L}_\text{v}$ for coarse affordance model training, \textit{c.f.} Supp.\ Mat.\ for details. $L_\text{g}$ and $L_\text{c}$ are used to supervise $\pi_\text{c}^\text{goal}$ and $\pi_\text{c}^\text{cont}$:

\begin{equation}
\begin{aligned}
\mathcal{L}_\text{g} = &\operatorname{BCE}(\hat{\mathbf{H}}_\text{g}, {\mathbf{H}}_\text{g})  +  \lambda_{\text{d}} \left\Vert\hat{{D}}_\text{g} - {D}_\text{g}\right\Vert_2^2 + \lambda_{\text{v}} \mathcal{L}_\text{v},
\end{aligned}
\label{eqn:loss_goal}
\end{equation}
where ${\mathbf{H}}_\text{g}$, ${D}_\text{g}$ are predicted by the goal predictor $\pi_\text{c}^\text{goal}$, $\lambda_{\text{d}}$ and $ \lambda_{\text{v}}$ are weighting factors.

\begin{equation}
\begin{aligned}
\mathcal{L}_\text{c} = \operatorname{BCE}(\hat{\mathbf{H}}_\text{c}, {\mathbf{H}}_\text{c})  +  \lambda_{\text{v}} \mathcal{L}_\text{v},
\end{aligned}
\label{eqn:loss_contact}
\end{equation}
where ${\mathbf{H}}_\text{c}$ is outputted by the contact predictor $\pi_\text{c}^\text{cont}$, $ \lambda_{\text{v}}$ is a weighting factor same as the one used in $\mathcal{L}_\text{g}$.

The fine affordance model $\pi_\text{f}$ is trained by supervising its output $\bar{\bm{\tau}}^{0}$ using the extracted trajectory $\hat{\bm{\tau}} \sim q(\bm{\tau})$:
\begin{equation}
\begin{aligned}
\mathcal{L}_\text{f} = \mathbb{E}_{\bm{\epsilon}, k}\left[ \left\Vert\hat{\bm{\tau}} - \bar{\bm{\tau}}^{0}\right\Vert_2^2\right],
\end{aligned}
\label{eqn:loss_fine}
\end{equation}
where $k \sim \mathcal{U}\{1,...,K\}$ is the diffusion step index, and $\bm{\epsilon} \sim \mathcal{N}(\mathbf{0}, \mathbf{I})$ is used to corrupt $\hat{\bm{\tau}}$ to obtain $\bm{\tau}^k$ inputted to $\pi_\text{f}$.
\section{Experiments}
\label{sec:experiments}
 
We aim to showcase the following aspects of our affordance model: (1) It significantly outperforms several strong baselines in zero-shot robot manipulation tasks. (2) Both coarse affordance prediction and test-time cost-guidance are crucial to guarantee its optimal performance. (3) It can enhance several downstream robot learning applications. (4) It can be seamlessly deployed across real robot systems. 

\begin{table*}[t]
\vspace{-0.12 in}

\centering
\small
\setlength{\tabcolsep}{8.5pt}
 \fontsize{7}{8}\selectfont
\begin{tabular}{@{}r|ccccccccccccc|c@{}}
\toprule
                 &  \textbf{\texttt{T01}}
                 &  \textbf{\texttt{T02}}       
                 &  \textbf{\texttt{T03}}
                 &  \textbf{\texttt{T04}}        
                 &  \textbf{\texttt{T05}}
                 &  \textbf{\texttt{T06}}           
                 &  \textbf{\texttt{T07}}
                 &  \textbf{\texttt{T08}}        
                 &  \textbf{\texttt{T09}}
                 &  \textbf{\texttt{T10}}        
                 &  \textbf{\texttt{T11}}
                 &  \textbf{\texttt{T12}}
                 &  \textbf{\texttt{T13}}
                & \textbf{\texttt{Avg.}}
                 \\
 
\cmidrule(lr){1-1}\cmidrule(lr){2-14} \cmidrule(lr){15-15}
GAPartNet \cite{geng2023gapartnet}  &   73.3    &    13.3        &        66.7     &  33.3      &      53.3     &     53.3      &     40.0       &       66.7      &       60.0     &         -       &    -      &    -       &    -       &    51.1    \\
Where2Act \cite{mo2021where2act}    &   86.7    &    \underline{53.3}       &        60.0     &  0.0          &     46.7      &     \underline{66.7}      &     60.0       &      \textbf{100.0}      &       53.3     &         -       &    -      &    -       &    -       &    58.5    \\
\cmidrule(lr){1-1}\cmidrule(lr){2-14} \cmidrule(lr){15-15}

Octo$^\text{*}$ \cite{team2024octo}         &   93.3    &    33.3        &        93.3     &  \textbf{66.7}      &     60.0      &     46.7      &     \underline{80.0}       &       \underline{80.0}      &       53.3     & \textbf{100.0} &   66.7    &    53.3    &    73.3    &    \underline{69.2}    \\
VRB$^\dagger$     \cite{bahl2023affordances}  &   \textbf{100.0}   &    20.0     &  \textbf{100.0}    &  46.7         &     \underline{66.7}      &     53.3     &       40.0     &       26.7   &      46.7      &\textbf{100.0}  &    60.0   &    46.7    &    60.0    &    59.0    \\
GFlow   \cite{yuan2024general}      &   \textbf{100.0}   &    33.3     & \textbf{100.0}    &  6.7          &     60.0      &     53.3     &       6.7      &       26.7   &      \underline{73.3}      &   93.3         &    \textbf{100.0}   &    \underline{60.0}    &    \underline{80.0}    &    61.0    \\
\textbf{Ours}                       &   \textbf{100.0}   &  \textbf{93.3}    & \textbf{100.0}    &  \textbf{66.7}       &    \textbf{80.0}      &     
\textbf{86.7}     &     \textbf{100.0}    &       66.7   &      \textbf{86.7}      &   \textbf{100.0}         &    \textbf{100.0}   &    \textbf{66.7}    &    \textbf{100.0}    &    \textbf{88.2}    \\

\bottomrule
\end{tabular}
\vspace{-0.1 in}
\caption{Quantitative results on 13 evaluated tasks evaluated on success rate (\%) in simulators. \textbf{\texttt{T01}}: Close hinge cabinet, \textbf{\texttt{T02}}: Close slide cabinet, \textbf{\texttt{T03}}: Close microwave,  \textbf{\texttt{T04}}: Open hinge cabinet, \textbf{\texttt{T05}}: Open microwave,  \textbf{\texttt{T06}}: Pull drawer, \textbf{\texttt{T07}}: Push drawer,  \textbf{\texttt{T08}}: Close dishwasher, \textbf{\texttt{T09}}: Open dishwasher,  \textbf{\texttt{T10}}: Pick up kettle, \textbf{\texttt{T11}}: Pick up can from clutter, \textbf{\texttt{T12}}: Pick up box from clutter, \textbf{\texttt{T13}}: Lift lid. $^\text{*}$: Fine-tuned on our extracted affordance data. $^\dagger$: Use strategy from \cite{kuang2024ram} to lift affordance to 3D. }
\label{tab:benchmark}.  
\end{table*}

\begin{table}[t]
\vspace{-0.3 in}

\centering
\small
\setlength{\tabcolsep}{1.5pt}
 \fontsize{7}{7}\selectfont
\begin{tabular}{@{}r|cccccc|c@{}}
\toprule
                 &  \textbf{\texttt{AT01}}
                 &  \textbf{\texttt{AT02}}       
                 &  \textbf{\texttt{AT03}}
                 &  \textbf{\texttt{AT04}}        
                 &  \textbf{\texttt{AT05}}
                 &  \textbf{\texttt{AT06}}
                & \textbf{\texttt{Avg.}}
                 \\
 
\cmidrule(lr){1-1} \cmidrule(lr){2-7} \cmidrule(lr){8-8}

\textbf{Ours [Full Model]}                       &   \textbf{93.3}   &  \textbf{66.7}    & \textbf{80.0}    &      86.7      &   \textbf{86.7}  & \textbf{100.0} & \textbf{85.6}       \\
\cmidrule(lr){1-8}  
w/o coarse goal. pred. \textbf{[V1]}               &  66.7            &  33.3    & 73.3     &     53.3                 &  60.0    & 60.0   &  57.8       \\
w/o multi-goal guide. \textbf{[V2]}                  &   80.0         &  40.0    & 60.0     &      \underline{93.3}                &   66.7   & \textbf{100.0}    & 73.3      \\
w/o contact-normal guide. \textbf{[V3]}              &  86.7    &  33.3    & \textbf{80.0}    &      \textbf{100.0}      &   60.0   & \textbf{100.0}  & 76.7     \\
w/o collision-avoid. guide. \textbf{[V4]}            &   \textbf{93.3}   &   \underline{60.0}    &   \textbf{80.0}    &  80.0         &   \underline{80.0}   &  73.3  &   \underline{77.8}  \\
w/o cost-informed heuristics \textbf{[V5]}            & 80.0   &   46.7    &  66.7   &   80.0       &    73.3 & \textbf{100.0}   &  74.5   \\

\bottomrule
\end{tabular}
\vspace{-0.1 in}
\caption{Ablation results on 6 selected tasks evaluated on success rate (\%). \textbf{\texttt{AT01}}: Close slide cabinet, \textbf{\texttt{AT02}}: Open hinge cabinet, \textbf{\texttt{AT03}}: Open microwave,  \textbf{\texttt{AT04}}: Pull drawer, \textbf{\texttt{AT05}}: Open dishwasher. \textbf{\texttt{AT06}}: Pick up can from clutter.}
\label{tab:ablation}.  
\vspace{-0.35 in}

\end{table}

\subsection{Experimental Setups}
\textbf{Simulator Environments.} 
We use IsaacGym~\cite{makoviychuk2021isaac} as our simulation platform for benchmarking, with environments developed based on \cite{li2024ag2manip}. We select 13 everyday household tasks from three widely-used benchmarks: FrankaKitchen \cite{gupta2019relay}, PartManip \cite{geng2023partmanip}, and ManiSkill \cite{gu2023maniskill2}. These tasks encompass action primitives such as opening, pushing, sliding, etc., and various objects, including cabinets, drawers, and kettles. Each task is evaluated from three different viewpoints. Each model generates five trajectories for each viewpoint, totaling 15 trials per task per model. Our evaluation protocol quantifies performance using the success rate (\%) commonly adopted in previous works, where a successful interaction is defined as causing the task object's degree of freedom (DoF) to exceed a pre-specified threshold without colliding with other objects in the scene.

\textbf{Baseline Models.} We compare our model against several representative baselines publicly available.
Specifically, GAPartNet \cite{geng2023gapartnet} and Where2Act \cite{mo2021where2act} are trained using virtual articulated assets collected (and interacted with) in simulators. Octo \cite{team2024octo} is pre-trained on a large-scale teleoperated dataset \cite{padalkar2023open} and further fine-tuned with our collected dataset. VRB \cite{bahl2023affordances}, GFlow \cite{yuan2024general} and our model are trained using human videos, while GFlow \cite{yuan2024general} can access the ground-truth depth, camera parameters, and object poses from \cite{Liu_2022_CVPR}. Hence, VRB \cite{bahl2023affordances} and ours operate in much more in-the-wild settings. We follow the strategy from \cite{kuang2024ram} to lift pixel-level trajectories from VRB \cite{bahl2023affordances} to 3D using object normal clusters as cues. We observed that baselines such as VRB~\cite{bahl2023affordances} and GFlow~\cite{yuan2024general} could not accurately infer contact regions. To ensure a fair comparison, we used our model instead to infer and standardize contact configurations. Hence, the benchmark focuses on predicting accurate interaction trajectories, which are more challenging than contact regions. We detail the strategy to obtain robot actions from our predicted affordance in Supp. Mat.

\subsection{Results and Discussions}
As shown in Table.~\ref{tab:benchmark}, our model achieves the best overall performance with an 88.2\% success rate across all tasks, surpassing the runner-up method by nearly 20\%. Among the baselines, VRB \cite{bahl2023affordances}, GAPartNet \cite{geng2023gapartnet}, and Where2Act \cite{mo2021where2act} exhibit inferior performance. We attribute this to their interaction policies being abstracted as directional vectors. While this approach works reasonably well for simpler tasks--such as pulling or pushing drawers along a straight line--it struggles with tasks requiring curved interactions, like opening a cabinet. The oversimplified motions tend to lead to gripper slip and subsequent task failures. Octo \cite{team2024octo} improves upon these models by approximately 10\%, achieving the second-best success rate of 69.2\%. This improvement is expected, given that Octo is pre-trained on a large-scale teleoperated dataset \cite{padalkar2023open} and demonstrates good generalization ability after being fine-tuned using our extracted dataset. However, the challenge of directly inferring complex actions from scene context may account for its lower performance than ours, as we quantitatively validate in Sec.~\ref{sec:ablation}. GFlow \cite{yuan2024general} suffers from erroneous trajectory scale predictions, leading to deteriorated performance.

The above results demonstrate that in-the-wild human videos can be a powerful data source for transferring manipulation skills to robots, provided that (pseudo) 3D interaction labels can be captured. However, the variance in scene contexts necessitates an effective affordance model capable of simultaneous high-level context understanding and low-level action planning that adapts to test-time constraints. 

\begin{figure*}[t]
	\centering 
        \vspace{-0.25in}
        \includegraphics[width=0.95\linewidth]{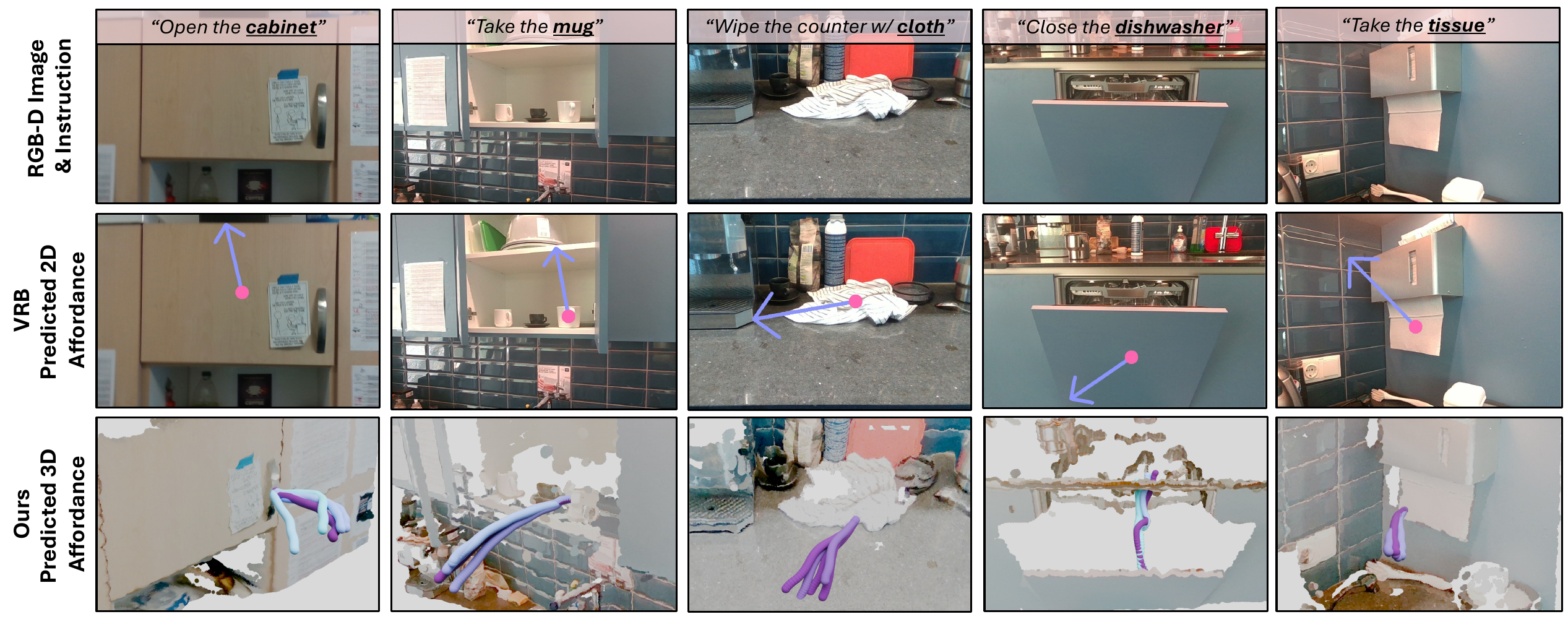} 
	\centering
        \vspace{-0.15in}
	\caption{\small Predicted affordance by VRB \cite{bahl2023affordances} and ours given instruction and RGB-D image. Though using the same RGB-only human videos for training, our framework predicts much more accurate contact points and interaction trajectories in 3D space directly, outperforming VRB \cite{bahl2023affordances} with ambiguous prediction in the pixel space. We visualize the top five affordance samples inferred by our model, where colors represent the final cost values; darker shades indicate lower costs and, therefore, a higher rank for the agent to execute.} 
     \label{fig:qualitative}
    \vspace{-0.2 in}
\end{figure*}

Notably, VRB \cite{bahl2023affordances} shares the same video data source \cite{damen2018scaling} as ours for affordance supervision. However, by fully exploiting 3D priors and employing our proposed affordance learning strategy, we significantly improved—boosting the task success rate by \textit{ca.}\ 30\%. Fig. \ref{fig:qualitative} demonstrates this enhancement using self-captured real-world data.

\subsection{Ablation Studies}
\label{sec:ablation}
As in Table.~\ref{tab:ablation}, we conducted detailed ablation experiments on a subset of manipulation tasks to assess the necessity of various important design choices in our affordance model. Specifically, \textbf{V1} is a variant without coarse goal prediction, hence also without goal conditioning and multi-goal guidance $\mathcal{J}_\text{goal}$. \textbf{V2}-\textbf{V4} are variants that generate trajectories without multi-goal guidance $\mathcal{J}_\text{goal}$, contact-normal guidance $\mathcal{J}_\text{contact}$ or collision-avoidance guidance $\mathcal{J}_\text{collide}$, respectively. \textbf{V5} randomly selects the generated trajectories instead of using the final guidance cost value as heuristics. 

\textbf{Impact of Coarse Affordance Prediction.} In \textbf{V1}, we observe a significant performance drop from 85.6\% to 57.8\%. This confirms that directly generating fine-grained actions from information-dense task observations is challenging. Inferring coarse goal configurations as \textit{affordance cues} from high-dimensional observation space simplifies the generation of accurate interaction trajectories.
\begin{figure}[t]
	\centering 
        \vspace{-0.35in}
        \includegraphics[width=0.95\linewidth]{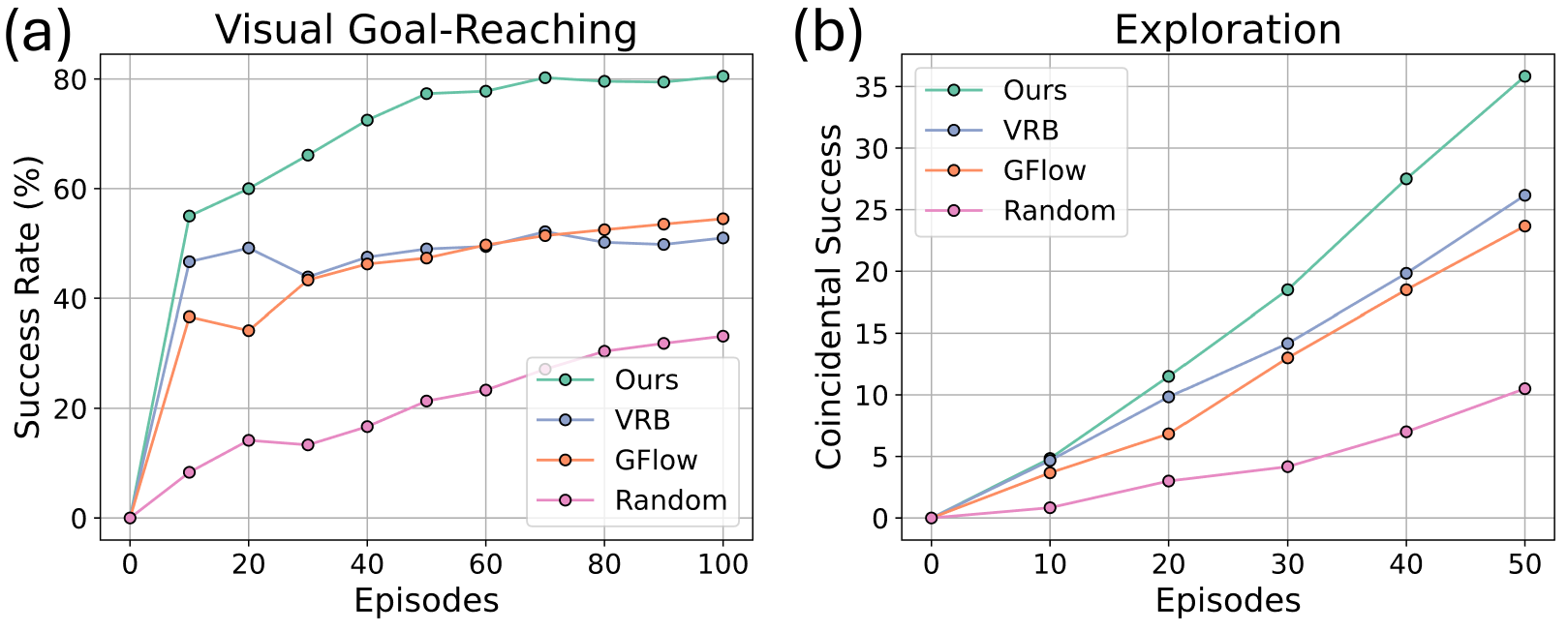} 
	\centering
        \vspace{-0.15in}
	\caption{\small (a) Average success rate for the visual goal-reaching task. (b) Average coincidental success for the exploration task.} 
         \label{fig:rlexps}
        \vspace{-0.25 in}
\end{figure}

\textbf{Impact of Cost Guidance.} 
Multi-goal guidance (\textbf{V2}) is the most crucial factor boosting overall performance by 12.3\%. Single-goal conditioning can mislead the generation process, whereas multi-goal conditioning through guidance helps correct accumulated errors from the coarse stage. Contact normal guidance (\textbf{V3}) is an intuitive hint for action generation, further improving performance. Collision avoidance guidance (\textbf{V4}) demonstrates its effectiveness, especially for the \textit{pick-up} task of portable objects (\textbf{\texttt{AT06}}) with a 26.7\% increase in success rate. These guidance terms, derived from test-time observations, enable more controllable trajectory generation under explicit constraints, enhancing the model's generalization toward unseen scenarios with new environments and embodiments. 

\textbf{Impact of Cost-informed Heuristics.} The performance drop of 11.1\% in \textbf{V5} underscores the importance of using the final guidance cost value as an intuitive criterion to select the optimal interaction plan.

\subsection{Robot Learning Applications}
\label{sec:robot_learn}
Following the robot learning paradigms introduced in \cite{bahl2023affordances}, we conducted several downstream application studies on the ablation tasks (\textbf{\texttt{AT01}}-\textbf{\texttt{AT06}}) to showcase the versatility of our affordance model as a strong prior.

\textbf{Visual Goal-Reaching.} The agent can additionally access an image of the object's desired configuration to enhance policy search supervision. Since our affordance model supports test-time guidance, we probabilistically replace the predicted goal points used in $\mathcal{J}_\text{goal}$ with those sampled from a buffer of successful trajectories in later episodes after collecting sufficient valid interactions. As shown in Fig.~\ref{fig:rlexps} (a), our method significantly outperforms other baselines trained with human videos, demonstrating faster convergence speeds and better overall performance. 

\textbf{Exploration.} The agent seeks to maximize the environment changes when interacting with the scenes. We employ the coincidental success metric as in \cite{bahl2023affordances}, i.e., the number of trajectories that bring the environment to the desired configurations without access to it. Similar to the goal-reaching tasks, we also use the previous successful trajectories to guide the sampling process of our affordance model. As illustrated in Fig.~\ref{fig:rlexps} (b), our method consistently demonstrates significant improvements over other baselines.

\subsection{Real Robot Experiments}
\begin{figure}[t]
	\centering 
        \includegraphics[width=0.95\linewidth]{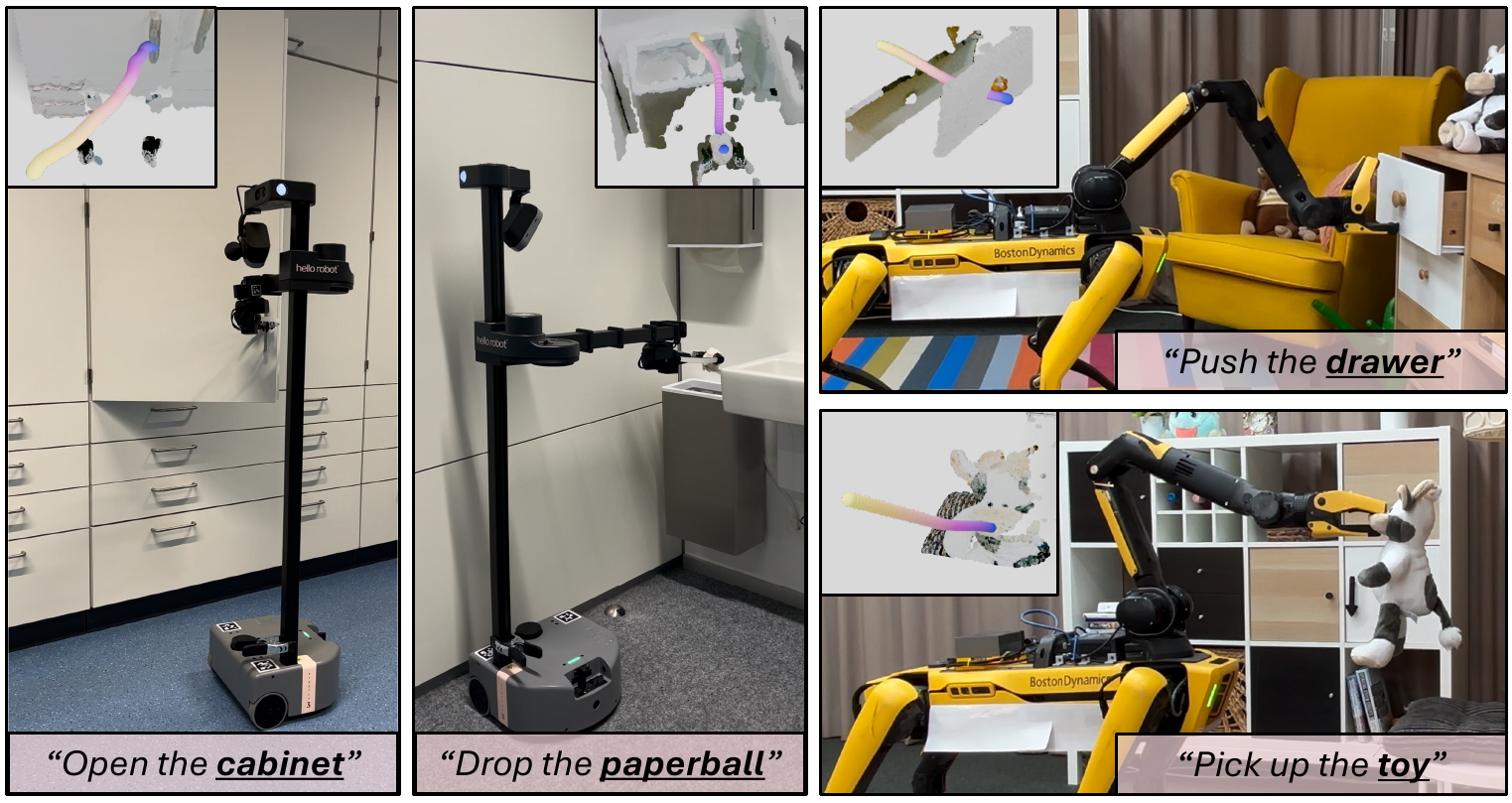} 
	\centering
        \vspace{-0.1in}
	\caption{\small Real-world robotic manipulation tasks with inferred affordance displayed in the top panels.} 
     \label{fig:qualitative_robot}
    \vspace{-0.15 in}
\end{figure}

We validate the efficacy of our framework on two real-world mobile robot platforms: the Hello Robot Stretch 3 and the Boston Dynamics Spot (\textit{c.f.}\ Fig.~\ref{fig:qualitative_robot}). Both robots are equipped with onboard RGB-D cameras for perception and receive language instructions for manipulation tasks. We test several household tasks within the robots’ physical capabilities, such as pushing a drawer, opening a cabinet, and taking a tissue, across three different human-suited environments. Overall, the robots achieved a success rate of 80.0\% over 55 trials, demonstrating the framework’s embodiment-agnostic nature and zero-shot transferability. Additional quantitative results are available in Supp.\ Mat.

\section{Conclusion}
\label{sec:conclusion}

In this work, we introduce \methodname, a scalable and effective framework that enables robots to learn manipulation skills directly from in-the-wild RGB-only human videos. 
\methodname demonstrates substantial generalization capabilities, outperforming existing methods by 20\% in success rate across 13 manipulation tasks in simulators in a zero-shot setting. Moreover, its embodiment-agnostic design allows for deployment across robot platforms, enabling successful executions of diverse household tasks in several real-world environments. The superior performance in downstream robot learning applications further underscores its versatility. One limitation of our framework is that the data quality is constrained by the accuracy of the depth foundation model and the SfM pipeline despite using the final optimization loss to filter low-quality labels. Given the method-agnostic nature of our data extraction pipeline, exploring recent learning-based SfM frameworks \cite{wang2024dust3r,duisterhof2024mast3r,zhang2024monst3r} could further enhance labels' quality. In future work, we plan to extract multi-modal affordance data using wearable devices, enabling robots to learn highly precise tasks like unscrewing caps, which currently remain challenging within our framework.

\section*{Acknowledgment}
\label{sec:acknowledgement}
This work was funded by TUM Georg Nemetschek Institute (GNI) via project SPAICR as well as gift funding from Google LLC. 
We thank Simon Schaefer, Simon Boche, Yannick Burkhardt, Leonard Freissmuth, Daoyi Gao, Yao Zhong for proofreading and fruitful discussions.

\clearpage
{
    \small
    \bibliographystyle{ieeenat_fullname}
    \bibliography{main}
}

\end{document}